\documentclass[11pt,a4paper]{article}
\usepackage[hyperref]{acl2021}
\usepackage{times}
\usepackage{latexsym}
\usepackage{array, tabularx, caption, boldline}
\usepackage{graphicx}
\usepackage{makecell}
\usepackage{float}

\usepackage{multirow}
\usepackage{amsmath}
\usepackage{amssymb}

\usepackage{microtype}

\aclfinalcopy 

\title{Uppsala NLP at SemEval-2021 Task 2: Multilingual Language Models for Fine-tuning and Feature Extraction in Word-in-Context Disambiguation}

\author{Huiling You, Xingran Zhu, and Sara Stymne \\
  Uppsala University \\
  Uppsala, Sweden \\
  \texttt{\{Huiling.You.7480,Xingran.Zhu.2781\}@student.uu.se}\\\texttt{sara.stymne@lingfil.uu.se}} 

\date{}

\begin{document}
\maketitle
\begin{abstract}

We describe the Uppsala NLP submission to SemEval-2021 Task 2 on multilingual and cross-lingual word-in-context disambiguation. We explore the usefulness of three pre-trained multilingual language models, XLM-RoBERTa (XLMR), Multilingual BERT (mBERT) and multilingual distilled BERT (mDistilBERT). We compare these three models in two setups, fine-tuning and as feature extractors. In the second case we also experiment with using dependency-based information. We find that fine-tuning is better than feature extraction. XLMR performs better than mBERT in the cross-lingual setting both with fine-tuning and feature extraction, whereas these two models give a similar performance in the multilingual setting. mDistilBERT performs poorly with fine-tuning but gives similar results to the other models when used as a feature extractor. We submitted our two best systems,  fine-tuned with XLMR and mBERT.
\end{abstract}

\section{Introduction}

SemEval-2021 Task 2: Multilingual and Cross-lingual Word-in-Context Disambiguation (MCL-WiC) \citep{martelli-etal-2021-mclwic} is an extension from WiC
\citep{pilehvar2019wic}, a shared task at the IJCAI-19 SemDeep workshop (SemDeep-5). WiC was proposed as a benchmark to evaluate context-sensitive word representations. The WiC dataset\footnote{\url{https://pilehvar.github.io/wic/.}} consists of a list of English sentence-pairs. Each sentence-pair
has a target word, and the task is to determine whether the target word is used in the same meaning or different meanings
in the two sentences, thus as a binary classification task. MCL-WiC extends WiC to multilingual and cross-lingual datasets,\footnote{\url{https://github.com/SapienzaNLP/mcl-wic}}
and covers 5 languages: Arabic, Chinese, English, French, and Russian. The MCL-WiC task is also framed
as a binary classification task: given a sentence-pair with a target word, either in the same language or in different languages,
the goal is to determine whether the target word is used in the same meaning or in different meanings.  Table \ref{tab:example} shows two example sentence pairs where the target word (\emph{mouse}) has either an `animal' or a `computer' sense.
In the \textbf{multilingual} setting, the two sentences are from the same language.  In the \textbf{cross-lingual} setting, the two sentences are from  different languages, English and one of the other four languages. Training data is only available for English--English, effectively leading to a zero-shot setting for the other languages.

\begin{table}
\centering
\begin{tabular}{|l|c|} \hline
\textbf{Example} & \textbf{Label} \\ \hline
The cat chases after the \emph{mouse}. & \multirow{2}{*}{F}  \\
Click the right \emph{mouse} button. & \\ \hline
The cat chases after the \emph{mouse}. & \multirow{3}{*}{T} \\
La \emph{souris} mange le fromage. & \\ 
(`The \emph{mouse} eats the cheese') & \\ \hline
\end{tabular}
\caption{Examples for multilingual (top) and cross-lingual (bottom) word-in-context disambiguation.} \label{tab:example}
\end{table}

Our main interest is to investigate the usefulness of pre-trained multilingual language models (LMs) in this MCL-WiC task, without resorting to sense inventories, dictionaries, or other resources. As our main method, we \textbf{fine-tune} the language models with a \emph{span classification head}. We also experiment with using the multilingual language models as \textbf{feature extractors}, extracting contextual embeddings for the target word. In this setting, we also add  information about syntactical dependency (i.e. head words and dependent words), with the intuition that it can contain relevant contextual information for disambiguation, as in Figure \ref{tab:example}, where the head words \emph{chases} and \emph{button} could help in disambiguating \emph{mouse}.  We compare three different LMs: XLM-RoBERTa (XLMR), multilingual BERT (mBERT) and multilingual distilled BERT (mDistilBERT).

We show that the fine-tuned models are stronger than any of the models based on feature extraction, by a large margin. XLMR is stronger than mBERT in the cross-lingual setting both with fine-tuning and feature extraction. mDistilBERT gives poor results with fine-tuning, but is competitive to the other LMs when used for feature extraction. Adding dependency syntax to our feature extraction method led to mixed results. We submitted our two strongest systems to the shared task, those fine-tuned with XLMR and mBERT. 

\section{Related Work}

In WiC at SemDeep-5, many participating systems capitalized on contextualized word representations. The LMMS (Language Modelling Makes Sense) system by \citet{loureiro2019language} used word embeddings from BERT, together with sense embeddings from WordNet 3.0 \citep{marciniak-2020-wordnet}. \citet{ansell-etal-2019-elmo} used the contextualized representations from ELMo \citep{peters2018deep} and trained a separate classification model. \citet{soler2019limsi} experimented with several contextualized representations and used cosine similarity to measure word similarities. \citet{wang2019superglue} included WiC as one of the tasks in the proposed SuperGLUE benchmark, with the approach of fine-tuning BERT. At the end of the WiC evaluation period, the best result was achieved by \citet{wang2019superglue} with an accuracy of 68.36\%, while human-level performance is 80\%, as provided by the dataset curators.

\citet{scarlini2020sensembert} recently proposed 
SensEmBERT\footnote{\url{http://sensembert.org/}}, a knowledge-based approach to sense embeddings for multiple languages. An important source for building SenseEmBERT is the contextualized representations from a pretrained language model. They  experimented with SensEmBERT on both English and multilingual word sense disambiguation (WSD) tasks, and showed that SensEmBERT is able to achieve state-of-the-art result on both English and multilingual WSD datasets.

\section{Multilingual Language Models}

\subsection{XLMR}
XLMR (XLM-RoBERTa) is a scaled cross-lingual sentence encoder \cite{conneau-etal-2020-unsupervised}, which is trained on 2.5T of data obtained from Common Crawl that covers more than 100 languages. XLMR has achieved state-of-the-art results on various cross-lingual NLP tasks.

\subsection{mBERT}
mBERT (multilingual BERT) is pre-trained on the largest  Wikipedias \cite{libovicky2019languageneutral}. It is a multilingual extension of BERT \cite{devlin-etal-2019-bert} that provides word and sentence representations for 104 languages, which has been shown to be capable of clustering polysemic words into distinct sense regions in the embedding space \cite{Wiedemann2019DoesBM}.

\subsection{mDistilBERT}
mDistilBERT (multilingual distilled BERT) is a light Transformer trained by distilling mBERT \cite{sanh2020distilbert}, which reduces the number of parameters in mBERT by 40\%, increases the speed by 60\%, and retains over 97\% of mBERT’s performance.

\subsection{Sub-word models}
XLMR, mBERT, and mDistilBERT all use sub-word models \cite{wu2016google,kudo2018sentencepiece}, so the target word is usually represented by several sub-tokens. For example, given ``qualify'' as target word, it will be represented by ``quali'' and ``fy'' in XLMR.  mBERT and mDistilBERT use a  WordPiece model with a vocabulary size of 119,447 and XMLR use a SentencePiece model with a vocabulary size of 250,002. In our work, when the target word is represented by multiple sub-words, we use the averaged embedding as feature vector for the target word.\footnote{We also explored summing sub-words, which gave similar results to averaging.}

\section{System Description}
We use the pre-trained language models in two different ways: for fine-tuning (Section \ref{sec:sys-ft}) and as feature extractors (Section \ref{sec:sys-target} - \ref{sec:sys-dep}).  Depending on whether feature transformation is involved, the features extracted can be further categorized into target word embeddings (Section \ref{sec:sys-target}) and dependency-based syntax-incorporated word embeddings (Section \ref{sec:sys-dep}). In the following sub-sections, we describe the three systems respectively.
Due to time constraints we did not use XLMR in the systems with feature extraction.

\subsection{Fine-Tuning}
\label{sec:sys-ft}

The fine-tuning setup follows the architecture designed by \citet{wang2019superglue},\footnote{The package for  SuperGLUE
tasks is available at \url{https://github.com/nyu-mll/jiant}} but extends to datasets in multiple languages. A
\emph{span classification head} is stacked on top of pre-trained language models, and attends only to the target words.
The \emph{span classification head} consists of a \emph{span attention extractor} and a classifier. The \emph{span attention extractor} is responsible
for extracting the span embeddings, namely the target words embeddings. First, the unnormalized attention score of each token
of the input document is computed. Span attention scores are the normalized scores of all tokens inside the span. Given the
attention distributions over spans, each span gets a weighted representation of the last-layer hidden states of either mBERT, mDistilBERT or XLMR.

\begin{figure}[t]
  \begin{center}
  \includegraphics[width=\textwidth,height=5cm,keepaspectratio=true]{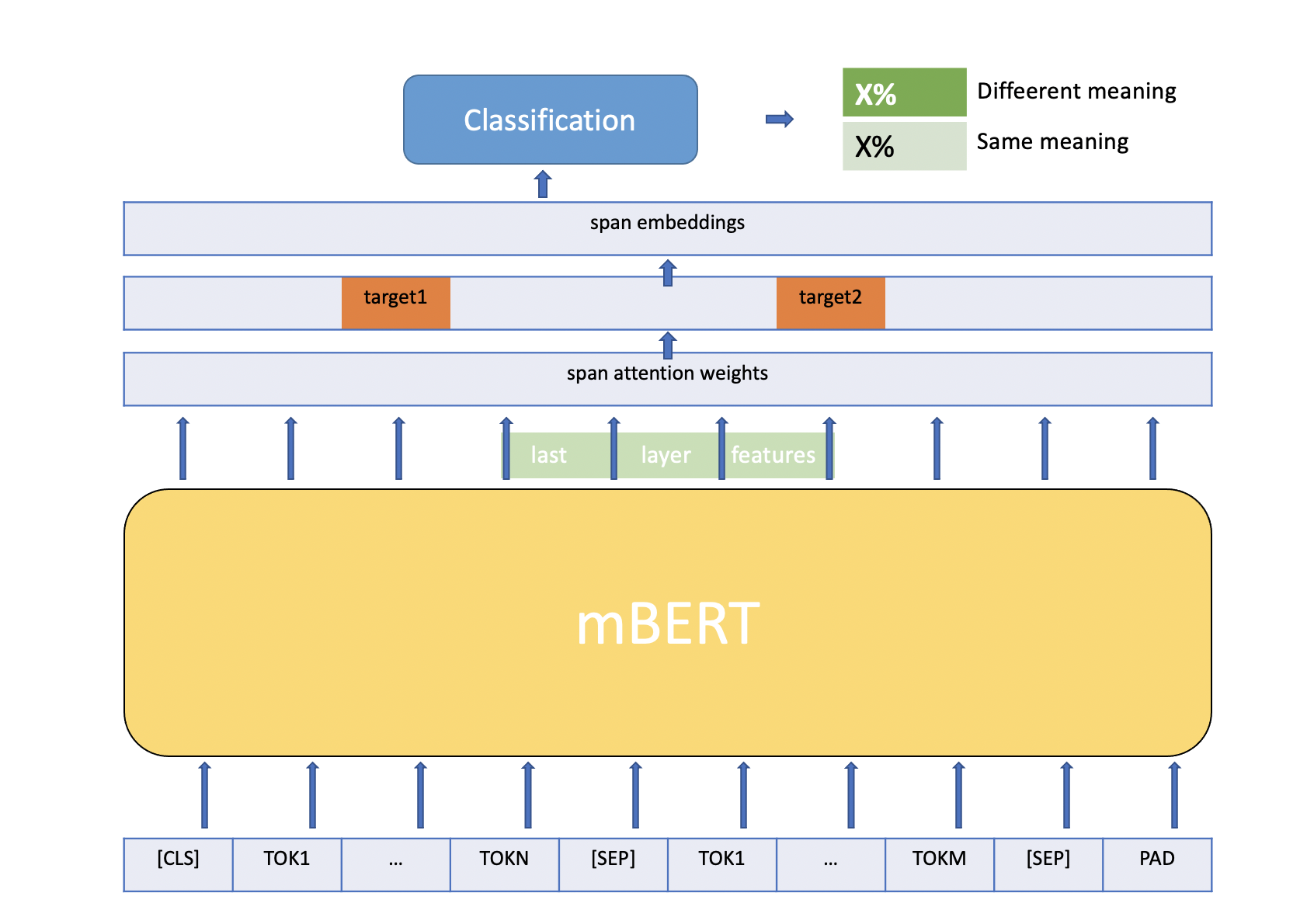}
  \caption{
      Model Structure of Fine-tuning mBERT
  }
  \label{fig:mbert fine-tune}
  \end{center}
\end{figure}

In this task, only the two target word spans will be returned, by masking out the rest of input.
The attended span embeddings are then passed to the classifier, a linear transformation layer, to produce the output logits,
which have a dimension of two, since there are only two labels (True or False).
Figure \ref{fig:mbert fine-tune} exemplifies the model structure when fine-tuning mBERT. The same structure also applies
to XLMR and mDistilBERT.

\subsection{Target Words Embeddings}
\label{sec:sys-target}

In this setup, the multilingual language models serve as pure feature extractors, to get target word embeddings from last-layer hidden states. The input sample of a sentence-pair will then be the concatenation of the pair of target word embeddings.

We feed the two sentences separately to the models, and concatenate the embeddings for the two target words.\footnote{We also experimented with concatenating the two sentences before feeding it to the LM, which gave slightly better results in some experiments. For consistency among all experiments we do not report these results.}
The extracted feature vectors are then fed to a classifier to perform the binary classification task. We experimented with two classifiers, logistic regression (LR) and a multi-layer perceptron (MLP). 

\subsection{Dependency-based Syntax-Incorporated Embeddings} 
\label{sec:sys-dep}
In this setup we ran a limited number of experiments. Only four languages (English, French, Chinese, and Russian)\footnote{The latest version of spaCy (3.0.0), which is the dependency parsing library used in this work, does not support dependency parsing for Arabic, thus we do not run experiments on Arabic in this setup.} and two pre-trained language models (mBERT and mDistilBERT) are explored.

The reasoning behind using syntax information to improve WiC classification results is as following. Given a pair of sentences, where the first sentence is ``The cat chases after the mouse'', and the second one is ``Click the right mouse button'', the target word \emph{mouse} has different head words: in the first sentence, the singular verb \emph{chases} is the head word, whereas in the second sentence, the noun \emph{button} is the head word. Since it is more natural for a real mouse (as a small rodent) to be \emph{chased} by its predators than to be related to a \emph{button}, while in contrast, it is more common for a computer mouse (as a hand-held pointing device) to have a \emph{button} than to be \emph{chased}, the head words therefore reveal information on different contexts of the target word. The same reasoning applies to dependent words as well.

First, each sentence is parsed using the spaCy  dependency parser,\footnote{\url{https://spacy.io/usage/linguistic-features\#dependency-parse}} from which we extract the target word, its head word, and its dependent word(s). 
Next, the sentence is passed to mBERT or mDistilBERT, and the corresponding target word embedding, head word embedding, and dependent word embedding(s) are retrieved, and concatenated. Note that if the target word has no head or dependent word, the \emph{null} token embedding\footnote{That is, simply feeding the word \emph{null} into mBERT/mDistilBERT and using the generated embedding directly.} is used instead; if the target word has more than one dependent word, all dependent word embeddings are summed element-wise.\footnote{We also explored averaging the dependent word embeddings, which gave equivalent results to summing.} Finally, the concatenated embeddings of two constituent sentences are further concatenated to form the sample feature vector of the sentence-pair, which is then fed to an MLP.

\begin{figure}[t]
  \begin{center}
  \includegraphics[width=\textwidth,height=9cm,keepaspectratio=true]{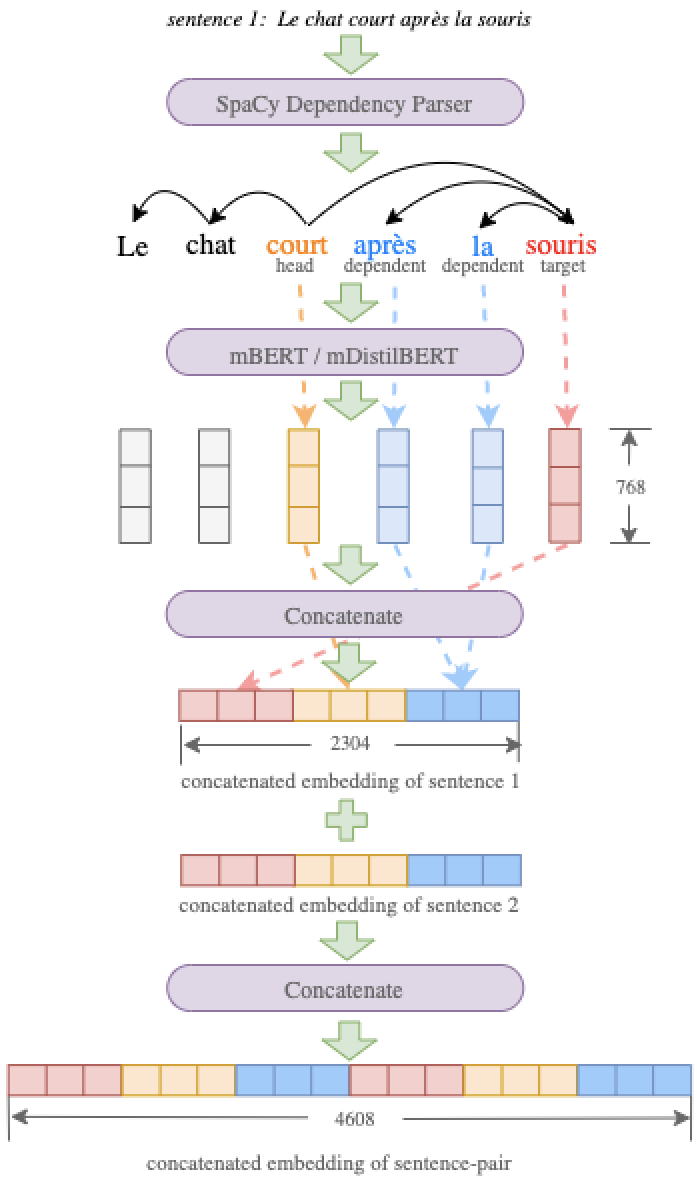}
  \caption{
      Construct a dependency-based syntax-incorporated embedding for a sentence-pair
  }
  \label{fig:syn_incop}
  \end{center}
\end{figure}

Figure \ref{fig:syn_incop} illustrates the process of constructing one such dependency-based syntax-incorporated embedding for a sentence-pair, of which the first sentence is \emph{Le chat court après la souris}. The default embedding size of mBERT/mDistilBERT is 768. The sizes of different concatenated embeddings are shown in Figure \ref{fig:syn_incop}. Again, we experimented with two classifiers, logistic regression and a multi-layer perceptron. 

\section{Experimental Setup}

\paragraph{Dataset} Only the datasets provided by SemEval-2021 Task 2 are used, see Table \ref{tab:dataset-table}. All systems are trained on the English set, the multilingual development sets are used during development, and the systems are tested on the multilingual and cross-lingual test sets.

\begin{table}[t]
  \centering
  \begin{tabular}{l|ccc}
  \Xhline{0.8pt}
  & Train & Dev & Test \\ \hline
   en-en    & 8000 & 500 & 1000 \\ \hline
    ar-ar & -- & 500 & 1000 \\  
    fr-fr & -- & 500 & 1000 \\  
    ru-ru & -- & 500 & 1000 \\  
    zh-zh & -- & 500 & 1000 \\  \hline
    en-ar & -- & -- & 1000 \\  
    en-fr & -- & -- & 1000 \\  
    en-ru & -- & -- & 1000 \\  
    en-zh & -- & -- & 1000 \\  \Xhline{0.8pt}
  \end{tabular}
  \caption{\label{tab:dataset-table} SemEval-2021 Task 2 Datasets. At development time, we only use half of the provided size (1000) of each dev set.}
  \end{table}

\paragraph{Fine-tuning} The three multilingual language models (mBERT, mDistilBERT, XLMR) are fine-tuned for three iterations, with batch size of 32, learning rate of 1e-5, and parameters optimized with
AdamW \citep{loshchilov2018decoupled}, provided by Huggingface's Transformers library \footnote{\url{https://huggingface.co/transformers/}}.

\paragraph{Logistic Regression} All logistic regression (referred to as ``LR'' in the following sections) models are trained for 150 iterations, with batch size of 32, learning rate of 0.0025 and parameters optimized with standard stochastic gradient descent (SGD).

\paragraph{MLP} All MLP models are 2-layer and follow the architecture suggested by \citet{du2019using}, outputing classification label based on the probability:
\begin{equation}
\mathbf{p} = \text{softmax}(L_2(\text{ReLU}(L_1(\mathbf{e}))))
\end{equation}
where $\mathbf{e}$ is in the input embedding, $L_i(\mathbf{x}) = \mathbf{W}_i \mathbf{x}+\mathbf{b}_i$ are fully-connected layers, $\mathbf{W}_1 \in \mathbb{R}^{H\times H}$ and $\mathbf{W}_2 \in \mathbb{R}^{2\times H}$ are layer parameter matrices, and $H$ is the input embedding size.
All MLP models are trained for maximum 200 iterations, with learning rate of 0.001 and parameters optimized with Adam ($\beta_1 = 0.9$, $\beta_2 = 0.999$) \cite{kingma2017adam}.

\paragraph{Language Model} We use the base version of all multilingual language models, with 12 layers, 12 attention heads, and hidden dimension of 768.
Due to time constraints we did not use XLMR in the  systems with feature extraction and an MLP.

\begin{table*}[ht]
\begin{scriptsize}
  \begin{center}
  \begin{tabular}{p{10mm}lccccccccc}
  \Xhline{0.8pt}
  & \bf System & \bf en-en & \bf zh-zh &  \bf fr-fr  & \bf ru-ru & \bf  ar-ar & \bf en-zh &  \bf en-fr & \bf en-ru & \bf en-ar \\ \hline
  
  \multirow{3}{*}{Fine-tune} & XLMR & \textbf{84.5\%}  & \textbf{78.3\%}  & 76.7\%  & 73.1\%  & 75.1\%  & \textbf{66.3\%}  & \textbf{70.9\%}  & \textbf{73.6\%}  & \textbf{65.2\%} \\
  
     & mBERT & 82.9\% & 76.2\% & \textbf{80.3\%} & \textbf{73.6\%} & \textbf{75.6\%} & 62.2\% & 66.3\% & 63.1\% & 59.4\% \\
  
  & mDistilBERT & 75.5\% & 68.0\% & 66.8\% & 64.8\% & 68.9\% & 51.8\% & 53.4\% & 51.9\% & 50.9\% \\
   \hline
  
  \multirow{7}{*}{\makecell[l]{Feature\\ Extractor}} &
     XLMR + LR & 53.9\%  & 55.4\% & 54.8\%  & 57.2\%  & 53.0\%  & 58.2\%  & 55.8\%  & 55.4\% & 54.7\%  \\
     & mBERT + LR & 53.4\% & 53.5\% &  49.7\%  & 51.7\%  & 53.1\%  &  52.0\% & 52.8\%  &  52.8\% & 51.1\% \\
  
  & mDistilBERT + LR & 55.7\% & 50.5\% & 52.6\% & 52.5\% & 51.9\% & 54.0\% & 52.5\% & 52.0\% & 51.6\% \\

   \cline{2-11}
  & mBERT +  MLP & 67.7\% & 51.4\% & 57.6\% & 54.2\% & 54.0\% & 47.4\% & 62.6\% & 55.6\% & 53.2\% \\
  & mDistilBERT + MLP & 66.6\% & 59.1\% & 59.8\% & 61.8\% & 56.0\% & 48.2\% & 63.2\% & 57.4\% & 52.3\% \\
  \cline{2-11}

    & mBERT + Syntax + MLP & 61.4\% & 52.7\% & 57.6\% & 57.0\% & -- & 53.4\% & 57.8\% & 55.6\% & -- \\
  & mDistilBERT + Syntax + MLP & 67.0\% & 56.6\% & 58.2\% & 57.6\% & -- & 54.0\% & 57.2\% & 56.2\% & -- \\
  \hline

  \Xhline{0.8pt}

  \end{tabular}
    \end{center}
  \end{scriptsize}
  
  \caption{\label{tab:results} System results on test sets. At task evaluation time, two fine-tuned systems were submitted, mBERT and XLMR; other systems were tested at post-evaluation time.}

  \end{table*}

\section{Results and Analysis}

The evaluation results on the test sets are shown in Table \ref{tab:results}. 
We can see that the fine-tuning approach is preferable to the feature extraction approach. All feature extraction variants fall behind the fine-tuned systems by a large margin. In many cases the systems based on feature extraction is just over chance performance (50\%), and in a few cases it is even below it. 

Among the fine-tuned systems, XLMR and mBERT give the best results, whereas mDistilBERT falls behind by quite a large margin in most cases, in several cases by more than 10 percentage points. The performance of mDistilBERT is especially weak in the cross-lingual setting. XLMR gives the best results for all cross-lingual language pairs, with an improvement over mBERT of 4.1--10.5 percentage points. The improvement is  largest for English--Russian. For the multilingual setting, the difference between mBERT and XLMR is smaller with at most 3.6 percentage points. XLMR gives the best score in two cases and mBERT in three cases.

Among the systems with feature extraction, the relative performance of the three sets of contextual embeddings differ from the fine-tuning. Here, mDistilBERT are competitive to the other two embeddings. We only use XLMR with LR, and again, we see that it gives the best performance in the cross-lingual setting among all systems with LR, just as with fine-tuning. In the multilingual setting, XLMR is also strong, having the best result for three out of five languages.
Compared to fine-tuning, mDistilBERT performs surprisingly well here. It is on par or better than mBERT in most cases across all settings.

Comparing the different architectures used with the feature extraction strategy, we see that using an MLP is preferable to LR, leading to large improvements in most cases. An exception is English--Chinese, where the MLP without syntax performs worse than LR. For  English--French on the other hand,  the MLP outperforms LR by around 10 percentage points, whereas we see small improvements for English--Russian. Finally, the addition of syntax leads to mixed results. For the English--Chinese system, we see large improvements, whereas we see the opposite for English--French.  For English--Russian as well as for all  multilingual systems, the differences are overall smaller. 

We also note that the performance is stronger for English--English than for the other languages in most settings. This is expected, since we only have English--English training data. A notable exception is for LR, where English--English performs considerably worse than in all other settings and is on par with the other languages in the same setting.
With fine-tuning we overall see stronger results in the multilingual setting, than in the cross-lingual setting, where we mix language pairs. We do not see this difference for our feature extraction systems, however.

\section{Conclusion and Future Work}

We have investigated the use of three large language models for multilingual and cross-lingual word-in-context disambiguation. We found that fine-tuning the language models is preferable to using them as feature extractors either for an MLP or for logistic regression. Trying to add dependency-based syntax information in the MLP gave mixed results. We also found that XLMR performed better than mBERT in the cross-lingual setting, both with fine-tuning and feature extraction, whereas the two models had a more similar performance in the multilingual setting. mDistilBERT did not perform well with fine-tuning, but was competitive to the other models in the feature extraction setting. We submitted our two best systems, fine-tuning with XLMR and mBERT to the shared task.

The fact that XLMR performs better than mBERT in the cross-lingual setting seems to indicate that it has a better representation of words across languages than mBERT and mDistilBERT. We think it would be worth investigating this hypothesis in more detail. XLMR and mBERT also use different sub-word models and another research direction is to explore the impact of this difference.  We would also like to investigate the effect of using representations from different layers of the pre-trained multilingual language models.

\bibliographystyle{acl_natbib}
\bibliography{mcl-wic}


\end{document}